# SCOUTING BY REWARD: VLM-TO-IRL-DRIVEN PLAYER SELECTION FOR ESPORTS[*]

Qing Yan[1], Wenyu Yang[2], Yufei Wang[1], Wenhao Ma[1], Linchong Hu[2], Yifei Jin[3], Anton Dahbura[1]

## 1. Introduction

Professional sports increasingly rely on data-driven methods for player evaluation, tactical analysis, and recruitment, leveraging dense event logs and, increasingly, player tracking data to quantify impact in ways that go beyond box-score statistics [1], [2]. Esports, and first-person shooters in particular, are in some sense even more data-rich: competitive titles such as Counter-Strike 2 log state–action trajectories at frame-level resolution and preserve full broadcast video of every round. Recent work has begun to treat esports as a structured performance domain with its own analytics questions, proposing index-like performance measures and machine-learning models that map detailed in-game behavior to outcomes [3], [4], [5]. Yet scouting workflows in top organizations still look strikingly traditional. Analysts watch hours of VODs, take qualitative notes on "feel" and "style," and then reconcile these impressions with metrics like rating or K/D. Existing data-driven systems largely aim to estimate overall skill or win contribution, rather than answering a question that is central to how clubs actually think about roster construction: "Who plays like our star?"

At the same time, inverse reinforcement learning (IRL) has emerged as a powerful way to infer latent reward functions from expert demonstrations, providing a principled framework for turning observed behavior into task objectives [6]. In traditional sports, IRL and related action-valuation methods have been applied to tracking and event data to derive context-aware value functions and player rankings that better reflect off-ball and defensive contributions [7], [8]. Classical Inverse Reinforcement Learning formulations focus on recovering a reward that explains approximately optimal behavior in a Markov decision process. We argue that this paradigm is particularly well-suited to the practical problem of style-based scouting in esports: a professional's play can be viewed as demonstrations of a policy, and the club's notion of "fit" can be represented as a reward function under which that policy is near-optimal.

In this paper, we propose a scouting system that learns pro-specific reward functions from logged gameplay and uses them to rank candidate players by fit to a given archetype. Concretely, we introduce a two-branch intake with a shared selection core: one branch encodes structured state–action trajectories from in-game telemetry, and a second branch encodes temporally aligned tactical commentary produced from broadcast footage. The branches are fused into a common representation that feeds a lightweight, GAIL-style (Generative Adversarial Imitation Learning; [9])

[*] The authors thank FNATIC Esports for their collaboration and for providing expert evaluators for this study.

[1] Johns Hopkins University
[2] University of Pennsylvania
[3] Cornell University

IRL objective. We validate the approach on Counter-Strike 2 clips collected in collaboration with a professional organization, comparing our IRL-based selector against human analysts and simpler baselines. Our results suggest that reward-centric, workflow-aware scouting can both match expert judgments and scale beyond what manual video review can reasonably cover.

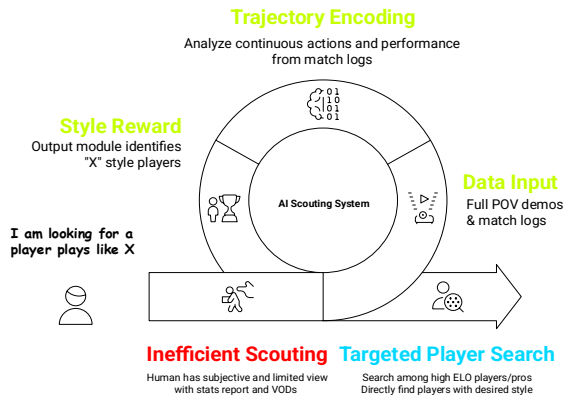

Figure 1: Why using AI, Why Scouting by Reward

## 1.1. Motivation

From a club's perspective, scouting is not only about finding high-skill players, but about identifying players whose decision-making and tendencies fit an existing system. Coaches and analysts talk about "playstyle," "archetypes," and whether a prospect "plays like" a current star or fills a specific tactical niche. Contemporary esports analytics tools mostly work at the level of aggregate statistics or generic performance indices [5]. They are useful for ranking players overall, but much less helpful for answering the practical question: "We need a support to assist our star just like Karrigan, who out there plays most similarly to him?"

This gap is especially stark in first-person shooters. The same headline metrics can mask very different patterns of risk-taking, tempo, and map control; two players with identical ratings may differ markedly in how they take duels, use utility, or support teammates—differences that recent interpretable machine-learning studies in Counter-Strike 2 make explicit when decomposing performance into multidimensional behavioral profiles [4]. Manual VOD review can, in principle, capture such nuances, but it does not scale to the volume of demos and streams and introduces inconsistency across scouts. At the same time, open circuits, regional leagues, and ranked ladders generate far more potentially promising players than any analyst team can systematically monitor.

Our work is motivated by the observation that most ingredients for style-aware scouting are already present. Modern titles supply dense state–action logs describing where and how players move and act, while broadcasts and commentary provide a human-centric view of the same scenes. Rather than hand-designing style features, we ask whether we can learn a reward function that embodies a specific professional's playstyle and then reuse that reward to screen new candidates. This reframes scouting "by eye" as scouting "by reward," with the potential to reduce analyst workload, increase consistency, and broaden the pool of players who can be fairly evaluated.

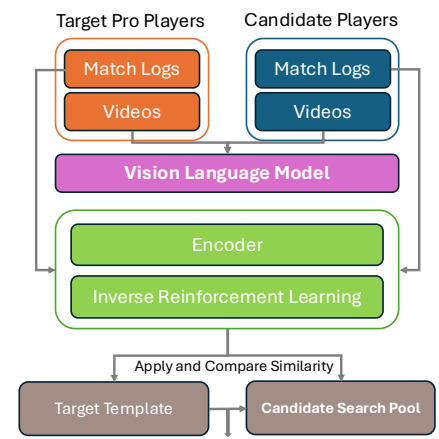

Figure 2: Scouting Model Architecture (ESIR)

## 1.2. Contributions

This paper makes three contributions. First, we cast style-based scouting in esports as an inverse reinforcement learning problem. Given demonstrations from a target professional, we learn a pro-specific reward over state–action trajectories to evaluate and rank candidate clips. To our knowledge, this is the first application of IRL to player-style similarity in esports; prior IRL work in sports has focused on valuing actions and ranking players within a single reward model aligned with league-wide notions of success [7], [8].

Second, we propose a practical scouting architecture tailored to data that esports clubs already have (Figure 2): a two-branch intake that fuses structured telemetry with temporally aligned tactical commentary from broadcast footage. Language produced from video—by a large vision–language model—serves primarily as an additional, flexible channel for describing game situations, while the core of the method lies in the shared representation and IRL-based selection core. This design keeps the dependence on any specific foundation model light and is compatible with ongoing advances in multimodal large language models [10].

Third, together with FNATIC Esports, we conduct an empirical study on Counter-Strike 2 clips collected with our professional club partner. We compare our IRL-based selector (ESIR) against human analysts, casual players, and simpler baselines that rely only on telemetry or generic similarity from pretrained models. We report accuracy on a closed-set identification task, agreement with expert rankings, and basic system performance, and we discuss how such a tool could sit inside a real scouting workflow. Together, these results suggest that learning to scout "by reward" is both feasible and practically relevant for esports organizations.

## 2. Related Work

Existing work touches our problem from three directions: (i) esports analytics and talent identification, (ii) inverse reinforcement learning and imitation-style evaluation in sports, and (iii) multimodal models that connect gameplay, structured data, and language. Our approach sits at their intersection: we borrow IRL ideas from sports analytics, attach them to the kind of telemetry used in esports player modeling, and use commentary embeddings only as a lightweight auxiliary signal rather than as the core contribution [6], [10], [11].

### 2.1. Esports Scouting and Player Analytics

Most recent esports analytics work focuses on predicting match outcomes or overall skill, not similarity to a specific professional's style. For example, Bahrololloomi et al. [3] build role-aware performance metrics for League of Legends and show that incorporating positional information improves win-prediction accuracy, while Chowdhury et al. [12] compare linear and ensemble models and highlight that most deployed tools still output a scalar win probability or rating. Birant and Birant [13] likewise frame esports analytics as multi-objective match-outcome prediction from structured statistics rather than style comparison between players.

Beyond match outcomes, growing work explores richer telemetry and role-specific features for modelling performance in MOBA and FPS titles but still treats skill as an absolute quantity (e.g., probability of winning, expected damage, or rank) rather than a relational notion of "plays like X" [3], [12]. Recent systems leverage multimodal data from streams and broadcasts—combining gameplay with chat or other side channels—to infer engagement, sentiment, or aggregate skill, but do not address club-defined archetypes or fit to a particular pro. In contrast, we formulate scouting around relative similarity: given a club-specified archetype, we learn a reward tailored to that professional's demonstrations and use it to rank candidates by compatibility rather than by a global "strength" score.

## 2.2. Inverse Reinforcement Learning in Imitation and Sports

Inverse reinforcement learning (IRL) and closely related imitation-learning methods infer latent objectives from expert demonstrations instead of hand-specifying rewards. Recent surveys group modern IRL algorithms into maximum-entropy, Bayesian, and adversarial families, emphasizing their use in high-dimensional robotics and control [6], [11]. Within this landscape, adversarial approaches such as GAIL-style methods are attractive for our setting: a discriminator learns to distinguish expert from non-expert trajectories, and its scores serve as a learned reward surrogate over state–action occupancy [9].

Sports analytics has begun to adopt IRL-style techniques on tracking and event data. Luo et al. [7] combine Q-function learning with IRL on millions of play-by-play events to derive player rankings under a single, league-wide reward model, while Rahimian and Toka [8] use batch IRL on soccer logs to infer interpretable offensive and defensive reward functions at the team level. In contrast, we learn separate, pro-specific reward surrogates with a GAIL-style discriminator for each professional and use them not for generic action valuation but for style-aware scouting and candidate ranking: the learned "reward" is calibrated to a particular star's demonstrations and reused to screen candidates for fit.

## 2.3. Vision-Language Models for Gameplay Analysis

Vision-language models (VLMs) have rapidly matured as general-purpose tools for fusing visual and textual information, with recent surveys outlining common architectures, training regimes, and failure modes. In games and sports, most work uses multimodal models to generate commentary or descriptions from logs or video, typically framing the problem as data-to-text. For example, Zhang et al.'s MOBA-E2C framework [14] generates Dota2 commentary from game metadata using a hybrid of rules and pretrained language models and constructs a large-scale esports commentary dataset. Recent systems for soccer and League of Legends likewise pair structured event records with commentary and use neural generators to produce broadcast-style text, and multimodal LLM/VLM architectures show that visual inputs can be combined with game data to improve the faithfulness and timing of descriptions.

Our use of VLMs is deliberately narrower. We do not tackle commentary generation or interactive game-playing; instead, we treat commentary—whether human or model-generated—as an additional, flexible channel for describing game situations that can be encoded alongside state–action telemetry. In this sense, we align with prior data-to-text and commentary work in viewing language as a high-level summary of play, but we depart from it by feeding commentary embeddings into an IRL-based selection core, where the main modeling novelty lies in the reward learning and scouting workflow rather than in the vision-language component itself.

# 3. Dataset and Evaluation Protocol

To investigate the feasibility of scouting via inverse reinforcement learning, we constructed a dataset that captures high-level professional play in Counter-Strike 2 (CS2). Our data pipeline transforms raw replay files into structured state–action trajectories suitable for reward learning, paired with ground-truth style annotations provided by domain experts.

## 3.1. Dataset of Gameplay Clips

We sourced our raw data from public .dem (demo) files available on HLTV.org and the FACEIT competitive platform. These files provide a lossless record of match history, capturing server-side state at a high tick rate. We filtered the corpus to focus on high-ELO competitive matches and professional tournament play to ensure the demonstrations reflect expert behavior. From these replay files, we extracted round-level telemetry using a custom parser based on the Valve Source 2 demo format. For each player in a replay, we generated a trajectory $\tau = \sum_N \{(s_t, a_t)\}_{t=0}^T$, where $s_t$ includes player location, equipment, health, and map control metadata, and $a_t$ represents low-level inputs (movement vectors, crosshair placement) and high-level events (utility usage, weapon firing) in total of N clip segments. The final dataset comprises approximately 30 clips of gameplay across 5 unique professional players, covering the active-duty map pool.

## 3.2. VLM-derived pseudo-commentary

In addition to structured telemetry, we also generate a lightweight semantic description for each gameplay clip. For every video segment, we query an off-the-shelf vision–language model with a constrained prompt that asks for a JSON trajectory of key actions every ~2 seconds (map, location, action, outcome, and impact labels). The model's output serves as our "pseudo-commentary": a machine-generated tactical description of the clip that is temporally aligned with the underlying state–action sequence. We treat these JSON events as text tokens and feed them into the commentary branch of our two-stream encoder, while the full prompt used for generation is included in Appendix A.[15]

## 3.3. Labels and Ground Truth Construction

To ensure the integrity of our expert archetypes, ground truth for the training set was established through a rigorous manual verification process. Our research team resolved the raw unique user identifiers (UUIDs)—specifically SteamIDs and FACEIT GUIDs—to confirmed real-world professional player identities, creating a noise-free repository of named expert demonstrations.

For the validation and test sets, we designed a blind "style fit" assessment to benchmark our model against human intuition. To prevent bias, team members manually sanitized the test clips, stripping away all identifying metadata such as in-game usernames, kill-feed notifications, and chat logs. This process left only visual information about player movement, crosshair placement, and mechanical execution, ensuring that assessment is based on gameplay style rather than reputation.

We adopted a granular 1–100 Likert scale for scoring. For a given query archetype (e.g., "Player A"), both human evaluators and our proposed model were tasked with assigning a similarity score (1–100) to a series of anonymous candidate clips. A score of 100 represents a perfect stylistic match to the anchor player, while lower scores indicate diverging decision-making or mechanics. This dual-annotation approach allows us to directly compare the model's internal valuation against the "eye test" performed by human analysts.

## 3.4. Evaluation Protocol

Our evaluation protocol focuses on the alignment between the automated IRL-based scores and the human-annotated ground truth. Since the model outputs a continuous reward or probability, we normalize these outputs to the 0–100 scale used by the human annotators.

We primarily evaluate Score Correlation and ranking consistency. We report the Pearson and Spearman correlation coefficients between the human-assigned fit scores and the model-predicted scores to quantify linear and monotonic relationships, respectively. A primary challenge in using human judgment as ground truth is individual rater bias; different analysts possess varying baselines for "similarity" and may utilize the 1–100 scale with different distributions (e.g., severity or leniency bias). To mitigate this, we apply Z-score normalization per annotator.

Furthermore, to validate the consistency of our human ground truth, we compute the Intraclass Correlation Coefficient (ICC) using a two-way mixed effects model. The ICC quantifies the degree of agreement between different analysts on the same set of clips. A high ICC value (>0.7) confirms that "style similarity" is an objective, observable property rather than a purely subjective impression, thereby validating the quality of the dataset used to benchmark the model.

# 4. Scouting Model Architecture and Baselines

Our proposed model is designed to learn a parameterized reward function $R_\theta(\tau)$ that assigns high values to state–action trajectories $\tau$ that resemble a target professional's style. The architecture consists of a dual-branch encoder for multimodal feature extraction, followed by a reward inference module trained via a GAIL-based IRL objective.

## 4.1. Multimodal Embedding Pipeline

The inputs to our model are twofold: structured telemetry from game logs and semantic descriptions from tactical commentary. As illustrated in Figure 2, both streams are fed into a behavior–style encoder that embeds discrete, multi-hot, and continuous features, applies a Transformer-based sequence model with mask-aware pooling to obtain a style vector, and uses a label encoder plus a GAIL-based IRL head to learn a reward $R_\theta(\tau)$ that scores trajectories by similarity to the target professional for candidate ranking.

## 4.2. IRL Reward Learning Objective

We approach the problem of style capture through the lens of Generative Adversarial Imitation Learning (GAIL). Unlike classical IRL, which requires solving a computationally expensive reinforcement learning loop to update a reward function at every step, GAIL directly extracts a policy and a reward signal by framing the problem as a minimax game between two competing networks: a Generator ($\pi_\theta$) and a Discriminator ($D_\phi$).

In our scouting context, the "Generator" acts as a Style Mimic, attempting to reproduce the target professional's decision-making, while the "Discriminator" acts as a Style Critic, learning to distinguish between the true professional's gameplay and the mimic's approximations.

We denote the expert (target professional) trajectories as $\tau_n$ and the generated trajectories as $\pi_n$. The Discriminator $D_\phi(s, a) \rightarrow [0, 1]$ is trained to output the probability that a given state-action pair (s, a) originates from the real professional data rather than the generator.

The objective function follows the standard Generative Adversarial Network (GAN) formulation:

$$\min_{\pi_\theta} \max_{D_\phi} V(\pi_\theta, D_\phi) = E_{\pi_E}[\log D_\phi(s, a)] + E_{\pi_\theta}[\log(1 - D_\phi(s, a))]$$

Once the adversarial training converges, the Discriminator $D_\phi$ implicitly encodes the "style" of the professional. We extract the Learned Style Reward $R_{style}$ directly from the discriminator's confusion. Following the GAIL formulation, we define the reward for a specific state-action pair as:

$$R_{style}(s, a) = -log(1 - D_\phi(s, a))$$

Intuitively, if the Discriminator is highly confident that a move belongs to the target pro (i.e., $D_\phi \approx 1$ ), the term $(1 - D_\phi)$ approaches 0, resulting in a high positive reward. Conversely, generic or uncharacteristic moves yield low rewards.

For the scouting phase, we freeze the parameters of the trained Discriminator. To evaluate a new candidate player, we feed their raw gameplay logs (states and actions) into this frozen Discriminator. The "Archetype Fit" score for a candidate trajectory $\tau_c$ is calculated as the mean accumulated reward over the episode:

$$Score(\tau_{candidate}) = \frac{1}{T} \sum_{t=0}^{T} -log(1 - D_\phi(s, a))$$

### 4.3. Explanations and User Interface

To make the system actionable for scouts, we developed a score interface that interprets the learned rewards.

**Similarity Score.** The raw reward output is normalized to a 0–100 scale (as described in Section 3) to provide an "Archetype Fit" metric.

**Temporal Similarity.** Shown in Figure 3, we highlighted moments in a round that contributed most to the high reward assignment to generate a temporal heatmap, answering the question: "Which specific decision made this player look like a pro player?"

### 4.4. Baseline Models

To benchmark the performance of our specialized IRL model against the general multimodal reasoning, we utilized three leading commercial Vision-Language Models (VLMs) as baselines: Gemini 3 Pro, GPT-5.1, and Claude 4.5.

These models were evaluated in a zero-shot setting to perform a direct similarity check. For each test clip, the VLMs were provided with the same visual input alongside a textual system prompt defining the stylistic attributes of the target professional. The models were instructed to analyze the visual mechanics and output a scalar similarity score (0–100) reflecting the degree of alignment between the observed gameplay and the target archetype.

This baseline comparison serves to isolate the value of the ESIR architecture. By contrasting our domain-specific reward function against these massive general-purpose engines, we test the hypothesis that identifying the fine-grained, sub-second mechanical signatures of professional play requires explicit reward learning, rather than high-level semantic pattern matching by VLMs.

# 5. Experiments and Results

To evaluate whether scouting "by reward" produces judgments comparable to professional analysts, we ran a controlled human–model study on Counter-Strike 2 gameplays under standardized conditions. We focused on five elite rifler/AWP archetypes on Dust2—m0NESY, b1t, donk, s1mple, and ZywOo—and trained five pro-specific IRL models on their POV demos. Each model was trained only on its assigned player's trajectories, using our GAIL-style discriminator as a learned style reward signal, and then frozen for downstream evaluation.

We then assembled a held-out pool of 150 short clips from Dust2 that neither the models nor the human evaluators had seen before. Clips were sanitized by stripping nicknames, HUD elements, radar, and kill-feed, and by muting audio so that both humans and models had to rely purely on visual state–action patterns rather than on caster cues or sound information. In-game cosmetics (e.g., weapon skins, glove patterns) could not be fully neutralized; we return to this potential source of bias in the limitations section.

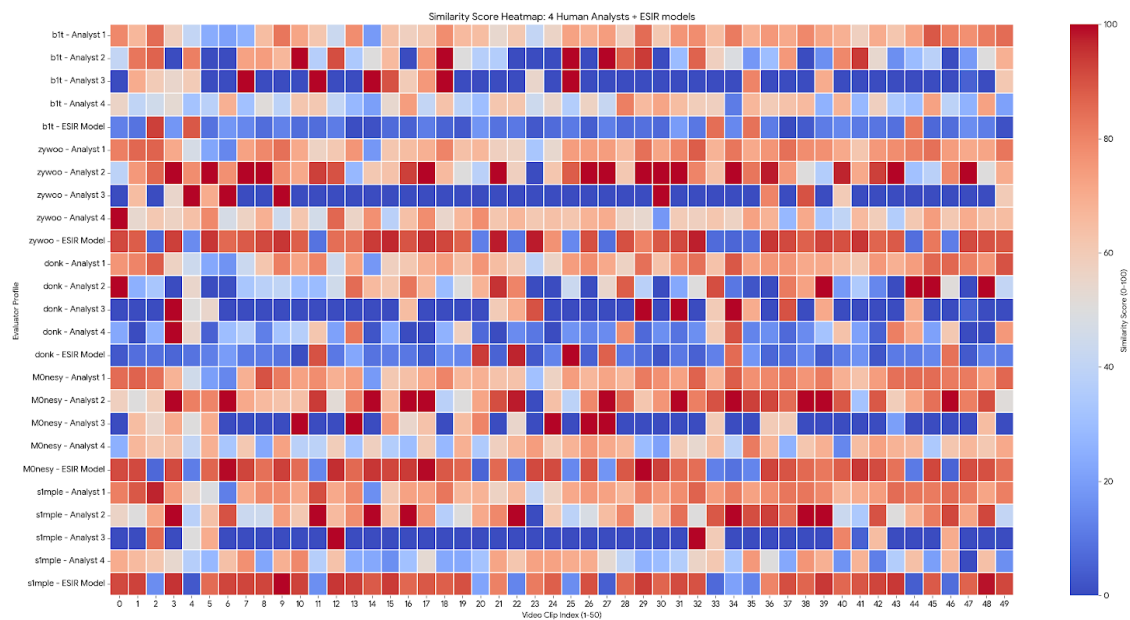

*Figure 3: Similarity Heatmap across 50 clips for Human Analysts and Model*

A professional coaching team from FNATIC Esports Team participated as expert raters. For each participant, we randomly sampled 50 clips from the pool (without replacement) and presented them in a web interface (Appendix B). For every clip, raters provided a 1–100 similarity score for each of the five anchor pros on independent sliders. In parallel, our five IRL selectors processed the same clips and produced a 5-dimensional vector of style-fit scores, one per clip, which we linearly rescaled to the same 1–100 range as human ratings. This yields paired human–model judgments for the same set of anonymous clips, enabling both identification and correlation-based evaluation.

## 5.1. Clip Identification Performance

By analyzing the comprehensive heatmap, we observe that human analysts exhibit distinct subjective baselines—manifesting as horizontal bands of consistently higher or lower intensity—which reveals individual biases in severity or leniency. While human evaluation is susceptible to cognitive fatigue and subconscious stereotypes regarding a player's reputation, the ESIR model functions as an invariant, data-driven anchor, shown in Figure 3. Consequently, the model not only validates consensus but also serves as a bias-mitigation tool; by flagging instances where a human score diverges significantly from the model's objective reward signal, organizations can identify and correct for "reputation drift" or personal preference, ensuring that scouting decisions remain grounded in the actual mechanics of gameplay rather than the scout's internal biases. While the correctness heatmap confirms that the ESIR model generally aligns with human consensus in identifying the true stylistic archetype, shown in Figure 4, it reveals a distinct divergence in confidence calibration. Human analysts frequently assign maximal similarity scores (approaching 100), reflecting a high degree of subjective certainty or 'all-or-nothing' judgment. In contrast, the model's learned reward function is markedly more conservative, rarely saturated at the upper bound even for correct identifications. This behavior is likely an artifact of the GAIL IRL objective, which penalizes overconfidence to maintain a robust policy distribution. Consequently, while the model matches human accuracy in determining the correct player, it avoids the 'false certainty' bias often observed in manual scouting, offering instead a nuanced, gradient-based measure of fit that implicitly acknowledges the stochastic nature of any single gameplay clip.

## 5.2. Ranking and Similarity Scores

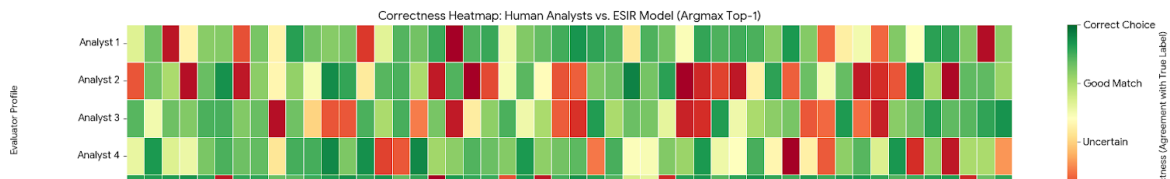

*Figure 4: Score Correctness Heatmap across 50 clips for Human Analysts and Model*

| Metric | Pearson Correlation (r) | Spearman Rank ($\rho$) | MAE(Z-Score Space) | Accuracy |
|---|---|---|---|---|
| **Human Analysts(Avg)** | 0.522 | 0.490 | 0.610 | 0.570 |
| **Our Model (ESIR)** | 0.447 | 0.393 | 0.701 | 0.840 |
| **Baseline Models** | -0.762 | 0.192 | 0.905 | 0.044 |

*Table 1: Human–Model Alignment and Accuracy on Style Identification*

To assess the continuous alignment between the evaluators and the ground truth scoring baseline, we computed the Pearson Correlation Coefficient (r), Spearman Ranking Correlation ($\rho$), and the Mean Absolute Error (MAE) in Z-score space [16], [17]. The comparative results for the average Human Analyst and the ESIR Model are presented in Table 1.

The Pearson Correlation Coefficient (r) measures the strength of the linear relationship between the evaluators and the ground truth, ranging from -1 to +1. In this study, Human Analysts achieved a higher correlation of r=0.522 compared to the Model's r=0.447. This disparity indicates that human scoring tends to track the consensus baseline in a more proportional, linear fashion; when the "true" quality of a clip increases, human scouts reliably increment their scores in moderate steps. The model's lower value of 0.447 does not necessarily imply poor performance but rather a non-linear scoring philosophy. Because the ESIR model acts as a discriminative agent— often jumping to high confidence scores for matches and near-zero for non-matches—it breaks the

"straight line" relationship that Pearson correlation rewards, prioritizing decisive classification over proportional tracking.

Moving beyond linearity, we evaluated the consistency of ordinal rankings using the Spearman Rank Correlation ($\rho$). This metric assesses whether the evaluator correctly sorts clips from "best fit" to "worst fit" regardless of the raw score values. Human Analysts outperformed the model with a $\rho = 0.490$ versus $\rho = 0.393$, suggesting that humans are more consistent at ranking the ambiguous "middle-of-the-pack" clips relative to the group consensus. While the model excels at identifying the Top-1 match (as seen in the correctness heatmap), its lower Spearman score reflects its binary nature; it tends to flatten the ranking of non-target clips to zero, losing the nuanced "2nd best" vs. "3rd best" distinctions that human analysts preserve in their rankings.

The cost of this discriminative behavior is ultimately quantified by the Mean Absolute Error (MAE), which measures the average magnitude of deviation in standardized Z-score space. The data reveals that Human Analysts maintained a lower error of 0.610, whereas the Model exhibited a higher deviation of 0.701. This lower error for humans reflects a "safety bias": analysts rarely commit to extreme outliers, preferring scores that cluster near the average, which mathematically minimize absolute error. Conversely, the model's higher MAE of 0.701 is a direct artifact of its "Argmax" confidence; by frequently committing to distinct high (>90) or low (<10) values, the model incurs a heavy statistical penalty whenever it disagrees with the consensus, effectively trading low-risk error minimization for high-confidence decision boundaries.

# 6. Discussion

Our study suggests that reward-based avatars of star players can already support practical scouting workflows but also highlights constraints in how they should be interpreted and deployed. We discuss implications for esports organizations and key limitations of our ESIR framework.

## 6.1. Implications for Esports Organizations

In our human–model comparison, the ESIR selector achieves a good–match accuracy of 0.840 on the five-way identity task, versus 0.570 for four professional analysts on the same 50 clips. Although narrow and dataset-specific, this gap indicates that a reward-based avatar of a star player can reliably match—and here exceed—expert judgement on "who plays like our pro" for short POV segments. In practical scouting terms, this means that when a club searches for "the next ZywOo" or "a donk-like rifler" among many candidates, an ESIR-style model is systematically more likely than a purely human pipeline to rank truly style-matched players near the top, including candidates whose box-score statistics look ordinary but whose trajectory-level behavior closely mirrors the target pro. For clubs, scouting therefore shifts from manual triage to continuous, model-driven search: instead of asking analysts to watch every promising ladder player, an ESIR-style model can pre-filter large candidate pools and surface a short list of trajectories that look highly compatible with a target archetype.

The broader context makes this attractive. Esports reach an audience of hundreds of millions, with Gen Z and young millennials dominating participation. As this cohort becomes the core of consumer spending, clubs and leagues will face pressure to professionalize recruiting and talent development at scale. Our per-pro reward models act as lightweight "digital twins" of star players: virtual replicas built from telemetry that let organizations test how thousands of candidate clips score under "the donk reward" or "the ZywOo reward," analogous to digital-twin systems used to model athlete performance in traditional sports [18], [19]. The same idea could extend beyond esports: avatars learned from trajectories could help scout creators in domains like music or short-form video by searching behavioral logs for "who behaves like this artist."

Although our vision–language baseline underperforms humans and ESIR on the identity task, recent work on VLM-based sports commentary and tactical analysis shows that multimodal models can turn raw video into structured tactical descriptions. As these models improve, organizations may deploy style-aware scouting without direct access to game telemetry, relying instead on broadcast video and pseudo-commentary as inputs to our reward-learning pipeline.

## 6.2. Limitation and Future Work

Several limitations temper these conclusions. Our user study had imperfect visual control, since some weapons and glove skins could not be fully neutralized, so participants may have used cosmetics as weak identity cues; future work should standardize inventories or use synthetic renders that remove cosmetics. We also evaluate a small, structured setting—five professionals on Dust2, 50 clips, one club—and optimize ESIR for top-1 agreement, whereas analysts appear better at ranking "near misses" and discussing second- and third-best fits; calibrating reward outputs for uncertainty and top-k recommendations is a natural extension. Finally, computational constraints restrict our vision–language component to generating pseudo-commentary rather than serving as a fully tuned, cross-game backbone; with larger-scale multimodal fine-tuning and more diverse titles, the same scouting-by-reward framework could generalize across games without per-title metadata parsers, further lowering deployment barriers for esports organizations.

# 7. Conclusion

By casting style-based scouting as an inverse reinforcement learning problem and validating ESIR against professional analysts, we showed that "scouting by reward" can already act as a stable, bias-mitigating avatar of star players, turning dense telemetry and lightweight commentary into archetype-fit scores and temporal explanations that scale far beyond what manual VOD review can cover. Building on this, we see our per-pro reward models as early digital twins of elite players: once a club can learn "the ZywOo reward" or "the donk reward" from trajectories, it is a short conceptual step to defining fully custom archetypes, composite "dream teammate" profiles, or even idealized "perfect player" templates and then searching massive ladders and open circuits for candidates who match those virtual personas. In the longer term, the same pipeline—multimodal intake, behavior–style encoding, and reward-based similarity—could extend across titles and into adjacent creator domains, where trajectories might be editing timelines, musical gestures, or content release patterns rather than crosshair movements. Our results suggest that ESIR is a useful first layer in such a system: a practical, workflow-aware scout that clubs can deploy today, and a foundation for future cross-game, cross-domain talent discovery engines that treat "who plays like our star?" as a general question about behavior in complex digital worlds, not just a one-off experiment on Dust2.

# Appendices

## Appendix A: VLM Prompt for Counter-Strike 2 Demo Parsing

```
You are a Counter-Strike 2 demo understanding model.

Given a video (and/or frames + audio) of a CS:GO round, your task is to ANALYZE the gameplay
and output a structured JSON object that summarizes key player actions every ~2 seconds. You
must output a trajectory through the whole video!!!!

### Output format (IMPORTANT)
You MUST output a SINGLE valid JSON object, with NO comments, NO trailing commas, and using
double quotes for all keys and string values.
The JSON MUST follow exactly this schema:

{
  "match_id": "..."
}

Do NOT include any comments in the JSON. Do NOT include any extra top-level keys.

### Allowed value pools (you MUST respect these)
MAP_POOL = ["de_mirage", "de_inferno"]
TEAM_POOL = ["CT", "T"]
ACTION_POOL = ["peek", "throw_grenade", "fire_weapon", "plant_bomb", "defuse_kit",
"hold_angle"]
WEAPON_POOL = ["ak47", "m4a4", "awp", "usp-s", "glock", "grenade", "smoke", "flash", "molly"]
LOCATION_POOL = {
  "de_mirage": ["mid", "A_site", "palace", "connector", "B_apps", "catwalk"],
  "de_inferno": ["banana", "B_site", "A_site", "long", "short", "apartments"]
}
OUTCOME_POOL = ["EnemySpotted", "Death", "EnemyDamaged", "FriendDamaged", "Assist"]
IMPACT_POOL = ["LossControl", "MapInformation", "CT_Depletion", "T_Advantage",
"ProjectileLoss"]

### Constraints
1. The "map" field MUST be one of MAP_POOL.
2. "team" MUST be one of TEAM_POOL.
3. "action" MUST be one of ACTION_POOL.
4. "weapon" elements MUST be from WEAPON_POOL.
5. "location" MUST be chosen from LOCATION_POOL[map] for the current map.
6. "outcome" MUST be a subset of OUTCOME_POOL (can be empty).
7. "impact" MUST be a subset of IMPACT_POOL (can be empty).
8. If some information is not visible, use consistent synthetic IDs (e.g., "player_1").

### Temporal sampling (every ~2 seconds)
You should conceptually scan the round timeline in ~2 second steps (0–2s, 2–4s, ...).
For each player, at each interval, if there is a notable action, create ONE trajectory entry
with:
- "timestamp": approximate time in seconds (e.g., 2.0).
- Best guess for "action", "location", and "result".
- If nothing important happens, you MAY skip adding an entry.

### Result object semantics
```

```
For each trajectory event:
- "outcome": list relevant outcomes (spotting, damaging, assisting).
- "impact": strategic impact from IMPACT_POOL.
- "targets": affected players.
- "weapon": main weapon used.
- "damage": integer (0-100).

### Final requirement
Your final answer MUST be ONLY the JSON object, with no explanations, no surrounding text, and
no markdown.
```

## Appendix B: The Web interface used to record the judgments of Pro esports analysts

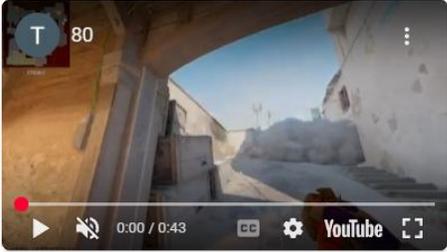